\documentclass[a4paper,conference]{IEEEtran}
\usepackage{ amssymb }
\usepackage{graphicx,color,psfrag}
\usepackage{multirow}
\usepackage{booktabs}
\usepackage{amsfonts} 
\usepackage{listings}
\usepackage{paralist}
\usepackage{amsmath}
\usepackage[hyphens]{url}
\usepackage{verbatim}
\usepackage{enumitem}
\usepackage{footnote}
\usepackage{threeparttable}
\usepackage{hyperref}
\usepackage{etoolbox}
\usepackage{multicol}
\usepackage{ tipa }
\usepackage{float}
\usepackage{mathtools}
\usepackage{varioref}
\usepackage{adjustbox}
\usepackage{rotating}
\usepackage{epstopdf}
\usepackage{bm}
\usepackage{caption}
\usepackage{subcaption}
\usepackage{breqn}
\usepackage{subfloat}

\ifCLASSOPTIONcompsoc
  \usepackage[nocompress]{cite}
\else
  \usepackage{cite}
\fi

\ifCLASSINFOpdf
 
\else
 
\fi

\begin{document}
\title{Semantic Scene Segmentation for Robotics Applications}
\author{\IEEEauthorblockN{Maria Tzelepi and Anastasios Tefas \\}
\IEEEauthorblockA{Department of Informatics\\
 Aristotle University of Thessaloniki\\
 Thessaloniki, Greece\\
 Email: $\{$mtzelepi,tefas$\}$@csd.auth.gr}}

\maketitle

\begin{abstract}
Semantic scene segmentation plays a critical role in a wide range of robotics applications, e.g., autonomous navigation. These applications are accompanied by specific computational restrictions, e.g., operation on low-power GPUs, at sufficient speed, and also for high-resolution input. Existing state-of-the-art segmentation models provide evaluation results under different setups and mainly considering high-power GPUs.
In this paper, we investigate the behavior of the most successful semantic scene segmentation models, in terms of deployment (inference) speed, under various setups (GPUs, input sizes, etc.) in the context of robotics applications. The target of this work is to provide a comparative study of current state-of-the-art segmentation models so as to select the most compliant with the robotics applications requirements.
\end{abstract}


\IEEEpeerreviewmaketitle

\section{Introduction}
Semantic scene segmentation describes the task of assigning a class label to each pixel of a given image, and thus it is also referred to as pixel-level classification. It constitutes a challenging step towards comprehensive scene understanding accompanied by many robotics applications \cite{meyer2017improved,maturana2018real,alonso2020mininet,milioto2018real}. Recent advances in Deep Learning (DL) provided effective models for addressing the problem of semantic scene segmentation \cite{long2015fully,badrinarayanan2017segnet}. However, most of the existing state-of-the-art DL segmentation models are usually computationally heavy, obstructing their deployment on robotics scenarios, where the deployment speed is critical. That is, in such scenarios, the segmentation models should be able to operate at sufficient speed on low-power GPUs, and also considering high-resolution input.

The performance reported by the existing segmentation methods refers to different setups (GPUs, input sizes, etc.), and mainly considering high-power GPUs.
Thus, in this paper we first discuss current state-of-the-art DL algorithms for semantic scene segmentation, and then we extensively evaluate them under different setups. That is, extensive evaluation of the existing segmentation models on different embedded and mobile platforms have been conducted, e.g., NVIDIA TX-2, AGX Xavier, and also for various input resolutions, ranging from lower ones, i.e., 512 $\times$ 512, to higher ones, i.e., 1024 $\times$ 2048. The objective of this work is to provide a comparative study of current  segmentation models considering the inherent computational restrictions in the context of robotics applications.

The remainder of the manuscript is structured as follows. Section \ref{s2} discusses recent efficient segmentation models, considering the deployment speed and the segmentation accuracy. The extensive evaluation of the most successful real-time segmentation models is provided in Section \ref{s3}, and finally conclusions are drawn in Section \ref{s4}.

\section{State-of-the-Art Segmentation Models}\label{s2}

During the recent years, several DL segmentation models have been proposed \cite{lateef2019survey,ulku2021survey}, achieving considerable performance on several image segmentation benchmark datasets, e.g., Cityscapes \cite{cordts2016cityscapes}, Camvid \cite{brostow2008segmentation}, COCO-Stuff \cite{caesar2018coco}, ADE20k \cite{zhou2016semantic}, PASCAL-S \cite{pascals}, and PASCAL-Context \cite{mottaghi_cvpr14}. 

More specifically, an efficient hierarchical multi-scale attention mechanism that allows the model to learn how to combine predictions from multiple inference scales is proposed in \cite{tao2020hierarchical} achieving state-of-the-art performance in terms of mean Intersection Over Union (mIOU) in the most widely used dataset for evaluating the performance of semantic segmentation methods, that is the Cityscapes test set. Subsequently, a method for improving the semantic segmentation performance by decoupling features into the body and the edge parts to handle inner object consistency and fine-grained boundaries jointly is proposed in \cite{li2020improving} achieving considerable segmentation performance. Furthermore, a method that proposes object-contextual representations to characterize a pixel by exploiting the representation of the corresponding object class is proposed in \cite{yuan2019object}, and achieves high performance in the aforementioned segmentation datasets. However, the aforementioned approaches focus on the segmentation accuracy achieving state-of-the-art performance, without addressing the issue of deployment speed. That is, they are computationally heavy, and thus inappropriate for robotics applications.

On the other hand, in the recent literature there are works that also focus on the deployment speed providing real-time segmentation models, mainly considering high-power GPUs (e.g., 1080Ti, 2080Ti). For example, a multi-resolution neural architecture search framework is proposed in \cite{chen2019fasterseg} achieving high segmentation accuracy and deployment speed at the same time. In addition, a fast segmentation model on high resolution input which proposes a learning to downsapmle module for computing low-level features for multiple resolution branches simultaneously is proposed in \cite{poudel2019fast} achieving high performance in terms of deployment speed, however comparatively lower performance in terms of segmentation accuracy, while the so-called Harmonic Densely Connected Network proposed in \cite{chao2019hardnet} achieves high performance in terms of accuracy however it runs at considerably fewer Frames Per Second (FPS) as compared to the competitive methods.

Subsequently, a Bilateral Segmentation Network (BiseNet) consisting of spatial path so as to preserve the spatial information and generate high resolution features, and a context path with a fast downsampling strategy so as to obtain sufficient receptive field is proposed in \cite{yu2018bisenet} achieving considerable performance both in terms of segmentation accuracy and deployment speed. Subsequently, the second version of the aforementioned method, proposing a detail branch so as to capture low-level details and a lightweight semantic branch so as to capture the high-level semantic context, as well as a booster training strategy for improving the segmentation performance without affecting the inference cost, is proposed in \cite{yu2020bisenet}. 
Furthermore, a method that proposes to re-design the commonly used residual layers so as to make them more efficient without affecting the learning performance is proposed in \cite{romera2017erfnet}, while a model namely ESPNet is proposed in \cite{mehta2018espnet} which based on an efficient spatial pyramid module, achieves high performance in terms of deployment speed with sufficient segmentation accuracy.

Subsequently, a method that employs an asymmetric encoder-decoder architecture where the encoder adopts the residual blocks and an attention pyramid network is employed in the decoder, is proposed in \cite{wang2019lednet}, while a lightweight segmentation model, namely LiteSeg, is proposed in \cite{emara2019liteseg}. The LiteSeg model, explores a new version of atrous spatial pyramid pooling, and achieves a considerable performance considering the accuracy-speed trade-off. Finally, a model where the encoder both encodes and generates the weights of the decoder is proposed in \cite{nirkin2020hyperseg}, while  a deep dual-resolution network, where a contextual information extractor is designed to enlarge effective receptive fields and fuse multi-scale context is proposed in  \cite{hong2021deep}, achieving also considerable performance considering the accuracy-speed trade-off.

However, the performance reported by the aforementioned segmentation methods refers to different setups (GPUs, input sizes, etc.), and mainly considering high-power GPUs. Thus, in this paper, we perform extensive experiments for evaluating the performance of the existing real-time segmentation models considering the deployment speed, with special emphasis on low-power GPUs, since we consider robotics applications.

\section{Experimental Evaluation}\label{s3}
\subsection{Experimental Setup}
The performance of the most successful real-time segmentation models previously presented is tested using the same setup on various GPUs, and also for various input sizes. Specifically, we perform experiments utilizing the BiseNetv1 \cite{yu2018bisenet}, BiseNetv2 \cite{yu2020bisenet}, HardNet \cite{chao2019hardnet}, LEDNet\cite{wang2019lednet}, ERFNet \cite{romera2017erfnet}, ESPNet \cite{mehta2018espnet}, and LiteSeg \cite{emara2019liteseg} models. The speed performance of the aforementioned models is tested on a RTX 2070, a Jetson TX2, a AGX Xavier, and a CPU. Experiments also conducted for various input sizes, that is 512 $\times$ 512, 512 $\times$ 1024, 1024 $\times$ 1024, and 1024 $\times$ 2048. The deployment speed is measured in terms of FPS. 

\subsection{Experimental Results}
\begin{figure*}[!h]
  \centering
\begin{subfigure}[b]{0.45\textwidth}
\includegraphics[width=\textwidth]{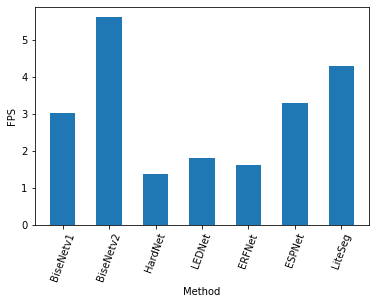}
\caption{Input Size: 1024$\times$1024.}
  \end{subfigure}
  \begin{subfigure}[b]{0.45\textwidth}
\includegraphics[width=\textwidth]{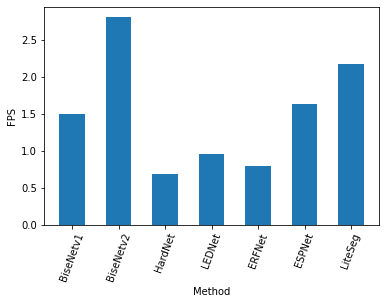}
\caption{Input Size: 1024$\times$2048.}
  \end{subfigure}
\caption{Semantic Segmentation Algorithms: Speed Evaluation (FPS) on TX2.}
  \label{fig:tx2-eval-seg2}
 \end{figure*}

\begin{figure*}[!h]
   \centering
\begin{subfigure}[b]{0.45\textwidth}
\includegraphics[width=\textwidth]{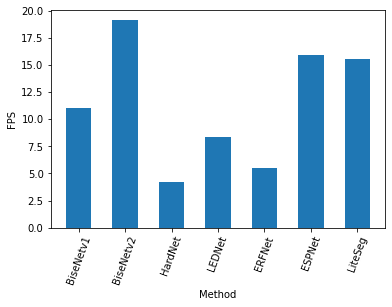}
\caption{Input Size: 1024$\times$1024.}
  \end{subfigure}
  \begin{subfigure}[b]{0.45\textwidth}
\includegraphics[width=\textwidth]{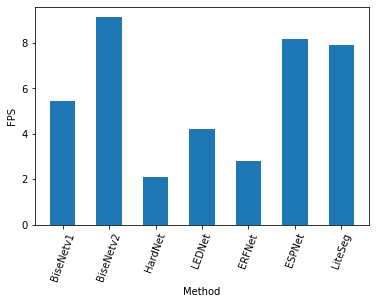}
\caption{Input Size: 1024$\times$2048.}
  \end{subfigure}
\caption{Semantic Segmentation Algorithms: Speed Evaluation (FPS) on AGX Xavier.}
  \label{fig:xavier-eval-seg2}
 \end{figure*}

Evaluation results for all the utilized segmentation models and on all the utilized GPUs,  considering the four different input resolutions are presented in Tables \ref{table:results-realtimeseg1}-\ref{table:results-realtimeseg4}. Furthermore, the comparison results considering the low-power GPUs, that is Jetson TX2 and AGX Xavier, and high-resolution input sizes, that is 1024$\times$1024 and 1024$\times$2048, are presented in Figs. \ref{fig:tx2-eval-seg2} and\ref{fig:xavier-eval-seg2}. 

\begin{figure}[!h]
\centering
\includegraphics[width=0.42\textwidth]{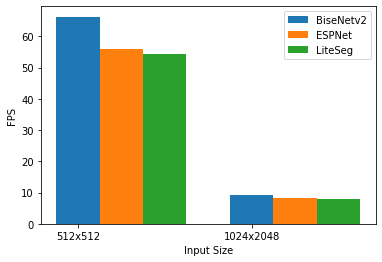}
\caption{FPS - Input Sizes, AGX Xavier.}
  \label{fig:xavier-fps-2inputsizes}
  \end{figure}

\begin{figure}[!h]
\centering
\includegraphics[width=0.5\textwidth]{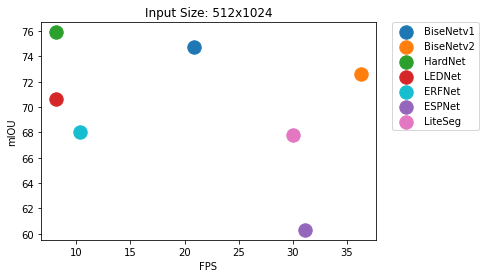}
\caption{mIOU - FPS: Input Size 512$\times$1024, AGX Xavier.}
  \label{fig:xavier-miou-fps}
  \end{figure}

As it demonstrated, BiseNetv2 runs faster on all the considered GPUs, except for the RTX 2070, and also for all the input resolutions. ESPNet, and LiteSeg achieve also considerable performance. Furthermore, it can be observed, as it also illustrated in Fig. \ref{fig:xavier-fps-2inputsizes} considering the AGX Xavier case, that for lower input resolution there are higher differences in terms of FPS among the most powerful considered models (i.e., BiseNetv2, ESPNet, and LiteSeg), while for higher input resolution there are slighter differences. Finally, in Fig. \ref{fig:xavier-miou-fps} the  segmentation accuracy-speed (in terms of mIOU and FPS, respectively) trade-off considering the AGX Xavier case for input size 512$\times$512 is illustrated, and it can be observed that the BiseNetv2 model is the most successful model considering the aforementioned trade-off, while considerable performance also achieve the LiteSeg and ESPNet models.

\begin{table}[!ht]
	\caption{Semantic Segmentation Algorithms: Speed Evaluation (FPS) Input Size: 512 $\times$ 512.}
	\centering
	\label{table:results-realtimeseg1}
	\begin{tabular}{|c|c|c|c|c|}
	\hline
	\textbf{Method} & \textbf{RTX 2070} & \textbf{TX2}& \textbf{AGX Xavier}& \textbf{CPU}\\ \hline
	BiseNetv1 (ResNet18) \cite{yu2018bisenet} & 170.43 & 11.25 & 39.06 & 8.32 \\ \hline
	BiseNetv2 \cite{yu2020bisenet} & 261.81 & \bf{21.53} & \bf{66.28} & \bf{15.69}\\ \hline
    HardNet \cite{chao2019hardnet} & 68.57 & 4.84 & 15.53 & 3.67\\ \hline
    LEDNet \cite{wang2019lednet} & 123.73 & 6.39 & 29.62 & 6.86\\ \hline
	ERFNet \cite{romera2017erfnet} & 124.01 & 5.70 & 19.71 & 6.86\\ \hline
    ESPNet \cite{mehta2018espnet} & \bf{265.08} & 12.04 & 55.79 & 7.98\\ \hline
    LiteSeg(MobileNet) \cite{emara2019liteseg} & 203.43 & 16.72 & 54.22 & 7.50\\ \hline
	\end{tabular}
\end{table}

\begin{table}[!ht]
	\caption{Semantic Segmentation Algorithms: Speed Evaluation (FPS) Input Size: 512 $\times$ 1024.}
	\centering
	\label{table:results-realtimeseg2}
	\begin{tabular}{|c|c|c|c|c|}
	\hline
	\textbf{Method} & \textbf{RTX 2070} & \textbf{TX2}& \textbf{AGX Xavier}& \textbf{CPU}\\ \hline
	BiseNetv1 (ResNet18) \cite{yu2018bisenet} & 93.84 & 5.92 & 20.83 & 4.10\\\hline
	BiseNetv2 \cite{yu2020bisenet} & 165.69 & \bf{10.94} & \bf{36.25} & \bf{6.84} \\\hline
    HardNet \cite{chao2019hardnet} & 39.80 & 2.60 & 8.14 & 1.86\\\hline
    LEDNet \cite{wang2019lednet} & 79.54 & 3.62 & 8.16 & 3.45\\\hline
	ERFNet \cite{romera2017erfnet} & 63.15 & 3.21 &  10.31& 3.07\\\hline
    ESPNet \cite{mehta2018espnet} & \bf{168.02} & 6.45 & 31.15 & 3.90\\\hline
    LiteSeg (MobileNet) \cite{emara2019liteseg} & 125.18 & 8.44 & 30.03 & 3.80\\\hline
	\end{tabular}
\end{table}

\begin{table}[!ht]
	\caption{Semantic Segmentation Algorithms: Speed Evaluation (FPS) Input Size: 1024 $\times$ 1024.}
	\centering
	\label{table:results-realtimeseg3}
	\begin{tabular}{|c|c|c|c|c|}
	\hline
	\textbf{Method} & \textbf{RTX 2070} & \textbf{TX2}& \textbf{AGX Xavier}& \textbf{CPU}\\ \hline
	BiseNetv1 (ResNet18) \cite{yu2018bisenet} & 49.11 & 3.03 & 11.02 & 2.02\\ \hline
	BiseNetv2 \cite{yu2020bisenet} & 85.55 & \bf{5.62} & \bf{19.13} & \bf{3.37}\\ \hline
    HardNet \cite{chao2019hardnet} & 21.29 & 1.37 & 4.18 & 0.93\\ \hline
    LEDNet \cite{wang2019lednet} & 43.06 & 1.80 & 8.38 & 1.69\\ \hline
	ERFNet \cite{romera2017erfnet} & 34.14 & 1.62 & 5.47 & 1.37\\ \hline
    ESPNet \cite{mehta2018espnet} &\bf{92.21}  & 3.29 & 15.96 & 1.85\\ \hline
    LiteSeg (MobileNet) \cite{emara2019liteseg} & 66.83 & 4.31 & 15.55 & 1.82\\ \hline
	\end{tabular}
\end{table}

\begin{table}[!ht]
	\caption{Semantic Segmentation Algorithms: Speed Evaluation (FPS) Input Size: 1024 $\times$ 2048.}
	\centering
	\label{table:results-realtimeseg4}
	\begin{tabular}{|c|c|c|c|c|}
	\hline
	\textbf{Method} & \textbf{RTX 2070} & \textbf{TX2}& \textbf{AGX Xavier}& \textbf{CPU}\\ \hline
	BiseNetv1 (ResNet18) \cite{yu2018bisenet} & 25.07 & 1.50 & 5.44 & 0.94\\\hline
	BiseNetv2 \cite{yu2020bisenet} & 44.56 & \bf{2.81} & \bf{9.13} & \bf{1.62}\\\hline
    HardNet \cite{chao2019hardnet} & 10.97 & 0.68 & 2.09 & 0.48\\\hline
    LEDNet \cite{wang2019lednet} & 21.97 & 0.96 & 4.20 & 0.83\\\hline
	ERFNet \cite{romera2017erfnet} & 17.61 & 0.80 & 2.81 & 0.66\\\hline
    ESPNet \cite{mehta2018espnet} & \bf{47.21} & 1.64 & 8.18 & 0.91\\\hline
    LiteSeg (MobileNet) \cite{emara2019liteseg} & 34.43 & 2.18 & 7.91 & 0.89\\ \hline
	\end{tabular}
\end{table}

\section{Conclusions}\label{s4}
In this paper, we first discussed existing state-of-the-art DL algorithms for semantic scene segmentation, and then we extensively evaluated them under different setups (on different embedded and mobile platforms, and also for various input resolutions). The objective of this work is to serve as a comparative study of current segmentation models considering the specific computational requirements in the context of robotics applications. Future research plans include the investigation of techniques for further improving the deployment speed of the most efficient models. 

\section*{Acknowledgment}
This project has received funding from the European Union's Horizon 2020 research and innovation programme under grant agreement No 871449 (OpenDR). This publication reflects the authors' views only. The European Commission is not responsible for any use that may be made of the information it contains.


\bibliographystyle{IEEEtran}
\bibliography{iisa_tzelepi}

\begin{thebibliography}{10}
\providecommand{\url}[1]{#1}
\csname url@samestyle\endcsname
\providecommand{\newblock}{\relax}
\providecommand{\bibinfo}[2]{#2}
\providecommand{\BIBentrySTDinterwordspacing}{\spaceskip=0pt\relax}
\providecommand{\BIBentryALTinterwordstretchfactor}{4}
\providecommand{\BIBentryALTinterwordspacing}{\spaceskip=\fontdimen2\font plus
\BIBentryALTinterwordstretchfactor\fontdimen3\font minus
  \fontdimen4\font\relax}
\providecommand{\BIBforeignlanguage}[2]{{%
\expandafter\ifx\csname l@#1\endcsname\relax
\typeout{** WARNING: IEEEtran.bst: No hyphenation pattern has been}%
\typeout{** loaded for the language `#1'. Using the pattern for}%
\typeout{** the default language instead.}%
\else
\language=\csname l@#1\endcsname
\fi
#2}}
\providecommand{\BIBdecl}{\relax}
\BIBdecl

\bibitem{meyer2017improved}
B.~J. Meyer and T.~Drummond, ``Improved semantic segmentation for robotic
  applications with hierarchical conditional random fields,'' in \emph{2017
  IEEE International Conference on Robotics and Automation (ICRA)}.\hskip 1em
  plus 0.5em minus 0.4em\relax IEEE, 2017, pp. 5258--5265.

\bibitem{maturana2018real}
D.~Maturana, P.-W. Chou, M.~Uenoyama, and S.~Scherer, ``Real-time semantic
  mapping for autonomous off-road navigation,'' in \emph{Field and Service
  Robotics}.\hskip 1em plus 0.5em minus 0.4em\relax Springer, 2018, pp.
  335--350.

\bibitem{alonso2020mininet}
I.~Alonso, L.~Riazuelo, and A.~C. Murillo, ``Mininet: An efficient semantic
  segmentation convnet for real-time robotic applications,'' \emph{IEEE
  Transactions on Robotics}, 2020.

\bibitem{milioto2018real}
A.~Milioto, P.~Lottes, and C.~Stachniss, ``Real-time semantic segmentation of
  crop and weed for precision agriculture robots leveraging background
  knowledge in cnns,'' in \emph{2018 IEEE international conference on robotics
  and automation (ICRA)}.\hskip 1em plus 0.5em minus 0.4em\relax IEEE, 2018,
  pp. 2229--2235.

\bibitem{long2015fully}
J.~Long, E.~Shelhamer, and T.~Darrell, ``Fully convolutional networks for
  semantic segmentation,'' in \emph{Proceedings of the IEEE conference on
  computer vision and pattern recognition}, 2015, pp. 3431--3440.

\bibitem{badrinarayanan2017segnet}
V.~Badrinarayanan, A.~Kendall, and R.~Cipolla, ``Segnet: A deep convolutional
  encoder-decoder architecture for image segmentation,'' \emph{IEEE
  transactions on pattern analysis and machine intelligence}, vol.~39, no.~12,
  pp. 2481--2495, 2017.

\bibitem{lateef2019survey}
F.~Lateef and Y.~Ruichek, ``Survey on semantic segmentation using deep learning
  techniques,'' \emph{Neurocomputing}, vol. 338, pp. 321--348, 2019.

\bibitem{ulku2021survey}
I.~Ulku and E.~Akagunduz, ``A survey on deep learning-based architectures for
  semantic segmentation on 2d images,'' 2021.

\bibitem{cordts2016cityscapes}
M.~Cordts, M.~Omran, S.~Ramos, T.~Rehfeld, M.~Enzweiler, R.~Benenson,
  U.~Franke, S.~Roth, and B.~Schiele, ``The cityscapes dataset for semantic
  urban scene understanding,'' in \emph{Proceedings of the IEEE conference on
  computer vision and pattern recognition}, 2016, pp. 3213--3223.

\bibitem{brostow2008segmentation}
G.~J. Brostow, J.~Shotton, J.~Fauqueur, and R.~Cipolla, ``Segmentation and
  recognition using structure from motion point clouds,'' in \emph{European
  conference on computer vision}.\hskip 1em plus 0.5em minus 0.4em\relax
  Springer, 2008, pp. 44--57.

\bibitem{caesar2018coco}
H.~Caesar, J.~Uijlings, and V.~Ferrari, ``Coco-stuff: Thing and stuff classes
  in context,'' in \emph{Proceedings of the IEEE Conference on Computer Vision
  and Pattern Recognition}, 2018, pp. 1209--1218.

\bibitem{zhou2016semantic}
B.~Zhou, H.~Zhao, X.~Puig, S.~Fidler, A.~Barriuso, and A.~Torralba, ``Semantic
  understanding of scenes through the ade20k dataset,'' \emph{arXiv preprint
  arXiv:1608.05442}, 2016.

\bibitem{pascals}
\BIBentryALTinterwordspacing
U.~CCVL, ``Pascal-s - the secrets of salient object segmentation dataset.''
  [Online]. Available: \url{http://cbi.gatech.edu/salobj/}
\BIBentrySTDinterwordspacing

\bibitem{mottaghi_cvpr14}
R.~Mottaghi, X.~Chen, X.~Liu, N.-G. Cho, S.-W. Lee, S.~Fidler, R.~Urtasun, and
  A.~Yuille, ``The role of context for object detection and semantic
  segmentation in the wild,'' in \emph{IEEE Conference on Computer Vision and
  Pattern Recognition (CVPR)}, 2014.

\bibitem{tao2020hierarchical}
A.~Tao, K.~Sapra, and B.~Catanzaro, ``Hierarchical multi-scale attention for
  semantic segmentation,'' \emph{arXiv preprint arXiv:2005.10821}, 2020.

\bibitem{li2020improving}
X.~Li, X.~Li, L.~Zhang, G.~Cheng, J.~Shi, Z.~Lin, S.~Tan, and Y.~Tong,
  ``Improving semantic segmentation via decoupled body and edge supervision,''
  \emph{arXiv preprint arXiv:2007.10035}, 2020.

\bibitem{yuan2019object}
Y.~Yuan, X.~Chen, and J.~Wang, ``Object-contextual representations for semantic
  segmentation,'' \emph{arXiv preprint arXiv:1909.11065}, 2019.

\bibitem{chen2019fasterseg}
W.~Chen, X.~Gong, X.~Liu, Q.~Zhang, Y.~Li, and Z.~Wang, ``Fasterseg: Searching
  for faster real-time semantic segmentation,'' \emph{arXiv preprint
  arXiv:1912.10917}, 2019.

\bibitem{poudel2019fast}
R.~P. Poudel, S.~Liwicki, and R.~Cipolla, ``Fast-scnn: Fast semantic
  segmentation network,'' \emph{arXiv preprint arXiv:1902.04502}, 2019.

\bibitem{chao2019hardnet}
P.~Chao, C.-Y. Kao, Y.-S. Ruan, C.-H. Huang, and Y.-L. Lin, ``Hardnet: A low
  memory traffic network,'' in \emph{Proceedings of the IEEE International
  Conference on Computer Vision}, 2019, pp. 3552--3561.

\bibitem{yu2018bisenet}
C.~Yu, J.~Wang, C.~Peng, C.~Gao, G.~Yu, and N.~Sang, ``Bisenet: Bilateral
  segmentation network for real-time semantic segmentation,'' in
  \emph{Proceedings of the European conference on computer vision (ECCV)},
  2018, pp. 325--341.

\bibitem{yu2020bisenet}
C.~Yu, C.~Gao, J.~Wang, G.~Yu, C.~Shen, and N.~Sang, ``Bisenet v2: Bilateral
  network with guided aggregation for real-time semantic segmentation,''
  \emph{arXiv preprint arXiv:2004.02147}, 2020.

\bibitem{romera2017erfnet}
E.~Romera, J.~M. Alvarez, L.~M. Bergasa, and R.~Arroyo, ``Erfnet: Efficient
  residual factorized convnet for real-time semantic segmentation,'' \emph{IEEE
  Transactions on Intelligent Transportation Systems}, vol.~19, no.~1, pp.
  263--272, 2017.

\bibitem{mehta2018espnet}
S.~Mehta, M.~Rastegari, A.~Caspi, L.~Shapiro, and H.~Hajishirzi, ``Espnet:
  Efficient spatial pyramid of dilated convolutions for semantic
  segmentation,'' in \emph{Proceedings of the european conference on computer
  vision (ECCV)}, 2018, pp. 552--568.

\bibitem{wang2019lednet}
Y.~Wang, Q.~Zhou, J.~Liu, J.~Xiong, G.~Gao, X.~Wu, and L.~J. Latecki, ``Lednet:
  A lightweight encoder-decoder network for real-time semantic segmentation,''
  in \emph{2019 IEEE International Conference on Image Processing
  (ICIP)}.\hskip 1em plus 0.5em minus 0.4em\relax IEEE, 2019, pp. 1860--1864.

\bibitem{emara2019liteseg}
T.~Emara, H.~E. Abd El~Munim, and H.~M. Abbas, ``Liteseg: A novel lightweight
  convnet for semantic segmentation,'' in \emph{2019 Digital Image Computing:
  Techniques and Applications (DICTA)}.\hskip 1em plus 0.5em minus 0.4em\relax
  IEEE, 2019, pp. 1--7.

\bibitem{nirkin2020hyperseg}
Y.~Nirkin, L.~Wolf, and T.~Hassner, ``Hyperseg: Patch-wise hypernetwork for
  real-time semantic segmentation,'' \emph{arXiv preprint arXiv:2012.11582},
  2020.

\bibitem{hong2021deep}
Y.~Hong, H.~Pan, W.~Sun, Y.~Jia \emph{et~al.}, ``Deep dual-resolution networks
  for real-time and accurate semantic segmentation of road scenes,''
  \emph{arXiv preprint arXiv:2101.06085}, 2021.

\end{thebibliography}

\end{document}